\documentclass[sigconf]{acmart}

\usepackage{amssymb}

\def\BibTeX{{\rm B\kern-.05em{\sc i\kern-.025em b}\kern-.08emT\kern-.1667em\lower.7ex\hbox{E}\kern-.125emX}}

\copyrightyear{2018}
\acmYear{2018}
\setcopyright{acmlicensed}
\acmConference[Woodstock '18]{Woodstock '18: ACM Symposium on Neural Gaze Detection}{June 03--05, 2018}{Woodstock, NY}
\acmBooktitle{Woodstock '18: ACM Symposium on Neural Gaze Detection, June 03--05, 2018, Woodstock, NY}
\acmPrice{15.00}
\acmDOI{10.1145/1122445.1122456}
\acmISBN{978-1-4503-9999-9/18/06}

\begin{document}

%
\title{Exploring Uncertainty Measures for\\Image-Caption Embedding-and-Retrieval Task}

%
\author{Kenta Hama}
\authornote{Both authors contributed equally to this research.}
\email{hamaken@ai.cs.kobe-u.ac.jp}
\affiliation{%
  \institution{Kobe University}
  \city{Kobe}
  \country{Japan}
}

\author{Takashi Matsubara}
\authornotemark[1]
\email{matsubara@phoenix.kobe-u.ac.jp}
\orcid{0000-0003-0642-4800}
\affiliation{%
  \institution{Kobe University}
  \city{Kobe}
  \country{Japan}
}

\author{Kuniaki Uehara}
\email{uehara@kobe-u.ac.jp}
\affiliation{%
  \institution{Kobe University}
  \city{Kobe}
  \country{Japan}
}

\author{Jianfei Cai}
\email{ASJFCai@ntu.edu.sg}
\affiliation{%
  \institution{Nanyang Technological University}
  \country{Singapore}
}

%
\renewcommand{\shortauthors}{Hama and Matsubara, et al.}

%
\begin{abstract}
With the wide development of black-box machine learning algorithms, particularly deep neural network (DNN), the practical demand for the reliability assessment is rapidly rising.
On the basis of the concept that ``Bayesian deep learning knows what it does not know,'' the uncertainty of DNN outputs has been investigated as a reliability measure for the classification and regression tasks.
However, in the image-caption retrieval task, well-known samples are not always easy-to-retrieve samples.
This study investigates two aspects of image-caption embedding-and-retrieval systems.
On one hand, we quantify feature uncertainty by considering image-caption embedding as a regression task, and use it for model averaging, which can improve the retrieval performance.
On the other hand, we further quantify posterior uncertainty by considering the retrieval as a classification task, and use it as a reliability measure, which can greatly improve the retrieval performance by rejecting uncertain queries.
The consistent performance of two uncertainty measures is observed with different datasets (MS COCO and Flickr30k), different deep learning architectures (dropout and batch normalization), and different similarity functions.
\end{abstract}

%
%

%

%

%
\maketitle

\begin{figure}[h]\centering
  \includegraphics[width=8.4cm]{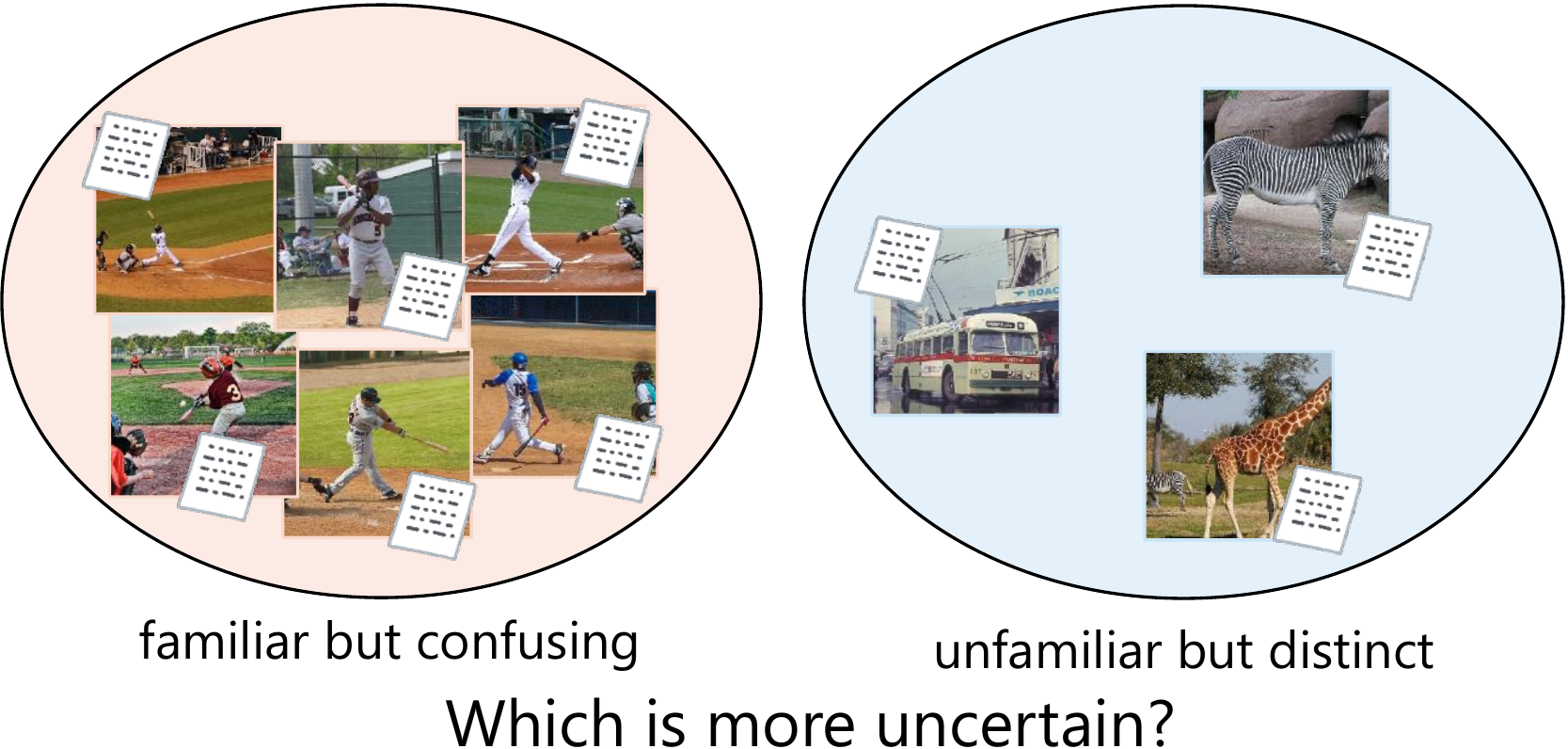}
  \vspace{-2mm}
  \caption{Conceptual diagram of uncertainty in image-caption embedding-and-retrieval task.}
  \label{fig:concept}
\end{figure}

\section{Introduction}
Recent advances in machine learning algorithms have enabled us to recognize and utilize various data modalities such as vision, natural language, and sound.
These algorithms, especially deep neural networks (DNNs), build black-box functions to make decisions in a data-driven manner.
While their accuracy is sufficient for many real-life applications, they have encountered safety issues, e.g., self-driving systems have injured pedestrians.
In addition to physical accidents, image-tagging systems and recommendation systems have also offended users by inappropriate suggestions.
These issues can be caused by insufficient training samples, dataset bias, and dataset shift~\cite{Leibig2016,Kendall2017a,Zhang2017o,Rohrbach2018}.
With a proper reliability assessment of their decisions, the applications can reduce the number of wrong decisions and ask for human intervention.
In this study, we quantify the uncertainty of image-caption embedding-and-retrieval systems as a first step toward assessing their reliability, as shown in Fig.~\ref{fig:concept}.

Several studies explicitly inferred a posterior distribution of a target instead of its point estimate and regarded the posterior entropy as aleatoric uncertainty~\cite{Kiureghian2009,Kendall2017a,Matsubara2018WCCI}.
The aleatoric uncertainty is expected to have a large value if a given sample is ambiguous and difficult to make a decision on.
However, a DNN classifier trained to minimize the empirical risk often outputs a posterior with almost zero entropy not depending on its reliability~\cite{Salimans2016b,Hein2019}.
Moreover, recent state-of-the-art approaches have employed regularized training objectives~\cite{Miyato2015,Zhang2017c,Takahashi2018a}, in which case, they no longer estimate true posteriors nor assess uncertainty.

Other approaches are based on Bayesian neural networks (BNNs)~\cite{Mackay1992,Hinton1993,Barber1997,Bishop2006}, in which the parameters are defined as random variables.
The output depends on a parameter (i.e., a model) drawn from its posterior.
The output averaged over these drawn models is often more accurate than a single output; this approach is called model averaging.
The variance of the output is expected to serve as an uncertainty measure called epistemic uncertainty~\cite{Kiureghian2009}.
Many recent DNNs can be BNNs because they employ stochastic components such as dropout~\cite{Srivastava2014} and batch normalization~\cite{Ioffe2015} in the training phase.
When these components are also used in the inference phase, they provide a different parameter and output for each run~\cite{Gal2016,Atanov2018}.
The epistemic uncertainty has been investigated for the classification task (including the segmentation task)~\cite{Gal2016,Kendall2017a,Kendall2017b,Leibig2016,Xiao2018a} and the regression task~\cite{Gal2016,Kendall2017a,Xiao2018a}, and has successfully detected misclassifications and excessive errors.
However, it still remains unclear how to define the epistemic uncertainty for other tasks such as the semantic embedding task.

This study focuses on the uncertainty from the two aspects of image-caption embedding-and-retrieval systems~\cite{Weston2010,Frome2013,Kiros2014,Karpathy2015,Zhang2017p,Gu2017,Faghri2018}.
The embedding task can be regarded as a regression task because its purpose is to arrange given samples in an embedded space~\cite{Taha2019}.
However, the performance of embedding has often been evaluated by the retrieval task, which is very similar to the classification task~\cite{Jain2016,Zhang2017p,Chen2019b}.

\textbf{Contributions.}
This study explores uncertainty measures for the embedding-and-retrieval task.
We demonstrate that the embedding-and-retrieval task can be regarded as a regression task and a classification task and that one has to evaluate an embedding-and-retrieval system in these aspects.
Section~\ref{sec:methods} proposes a new approach to estimate the posterior distributions that a given query is associated with a target sample even though the objective function is the rank loss~\cite{Weston2010,Frome2013,Kiros2014,Karpathy2015,Zhang2017p,Faghri2018}.
Then, we propose feature and posterior uncertainties from viewpoints of regression and classification tasks, respectively.
In Section~\ref{sec:results}, we demonstrate that, for improving the performance of the retrieval task, model averaging based on these two uncertainties works similarly well.
For assessing the reliability, the posterior uncertainty provides a better reliability measure for the retrieval task than the feature uncertainty.
These tendencies are common for different datasets, DNN architectures, and similarity functions.
Also, the posterior uncertainty quantifies the dataset shift, whereas the feature uncertainty cannot do this.
A qualitative comparison of these uncertainties reveals that the tendencies are caused by the biases in multi-modal datasets; a dataset containing many similar samples makes the DNN familiar with and certain about these samples, but the same dataset provides the confusing task of retrieving a desired sample from many similar samples, as exemplified in Fig.~\ref{fig:concept}.
All results are based on the MS COCO~\cite{Lin2014}, Flickr30k~\cite{Young2014}, and RecipeQA~\cite{Yagcioglu2018} datasets.

\section{Related Works}
\subsection{Bayesian Neural Networks}
A typical DNN is a map from an input $x$ to an output $y$ parameterized by $w$, which is expressed as $y=DNN(x,w)$, hereafter.
Given a dataset $D=\{(x_i,t_i)\}_{i=1}^N$ for a classification task, the target $t_i$ is the class to which the input $x_i$ belongs, and the output $y_i$ is typically an estimate of its posterior probability $y_i=p(t_i|x_i,w)$.
For a regression task, the output $y_i$ is a point estimate of the target $t_i$ or its posterior distribution $p(t_i|x_i,w)$, expressed using the reparameterization trick~\cite{Kingma2014,Kendall2017b,Kendall2017a}.

A BNN is a neural network whose parameter $w$ is estimated as a posterior distribution $p(w|D)$ given a dataset $D$.
Because the true posterior $p(w|D)$ is intractable, many approximation methods have been proposed~\cite{Mackay1992,Hinton1993,Barber1997,Bishop2006}.
Recent studies have revealed that the optimization of a DNN using stochastic components such as dropout~\cite{Srivastava2014} and batch normalization~\cite{Ioffe2015} provides an approximated posterior $q(w|D)$ of the parameter $w$~\cite{Gal2016,Atanov2018}.
The original proposers supposed to average the parameter $w$ over the approximated posterior $q(w|D)$ in the inference phase.
Specifically, the output is $y=DNN(x,\bar w)$ for $\bar w=\mathbb{E}_{q(w|D)}[w]$, and this approach is called \emph{weight averaging}~\cite{Kendall2017b,Kendall2017a}.
However, one can draw a new instance of the parameter $w$ from the posterior $q(w|D)$ even in the inference phase.
Each drawn instance of the parameter $w$ provides a single model $p(t|x,w)$.
Drawing multiple models, one can obtain a more accurate posterior by averaging the posteriors as
\begin{equation}
  \begin{split}
    y&=p(t|x,D)\\
    &=\int_w p(w|D)p(t|x,w)\\
    &\approx\int_w q(w|D)p(t|x,w).
  \end{split}\label{eq:averaging}
\end{equation}
This approach is called \emph{model averaging}~\cite{Bishop2006,Kendall2017b}.
Because the integral of a DNN is intractable, the expectation over the posterior $q(w|D)$ is approximated using Monte Carlo sampling of $L$ models.
Many previous studies have confirmed that the model averaging improves the classification accuracy~\cite{Kendall2017b,Kendall2017a,Gal2016,Atanov2018,Xiao2018a}.
The regression task often employs non-probabilistic measures such as the mean squared error and $L_1$ distance, and the rigorous model averaging does not always improve these measures.
Instead, the output $y$ is simply averaged over the weight posterior $q(w|D)$~\cite{Gal2016,Kendall2017a,Xiao2018a,Taha2019} as
\begin{equation}
  \begin{split}
    y=\mathbb E_{q(w|D)}[DNN(x,w)]
  \end{split}\label{eq:regression_averaging}
\end{equation}

\subsection{Uncertainty Quantification}
In an exact Bayesian inference, the parameter $w$ first follows a prior $p(w)$ and then is gradually specified as the number of observed samples increases.
The posterior distribution $p(t|x_i,D)$ also has a large variance at the beginning and then approaches the target $t_i$~\cite{Watanabe2010b}.
Similarly, the BNN is trained to provide an output $y_i$ robust to the stochastic behavior of dropout and batch normalization.
Given unfamiliar inputs such as samples from a different dataset or domain, the outputs still suffer from the stochastic behavior and have large variances~\cite{Kendall2017a,Leibig2016}.
This behavior is similar to that of a Gaussian process~\cite{Gal2016}.
The output variance $Var_{q(w|D)}[y]$ is an uncertainty measure called epistemic uncertainty, and it is available for assessing the reliability of decision making.
This usage is based on the assumption that a decision on an unfamiliar sample is unreliable.
For the classification task, an alternative is the mutual information $I(t;w|x,D)$ between the class posterior and the parameter $w$~\cite{Smith2018a}.
The variance of the class posterior $p(t|x,D)$ can be considered as a first-order approximation of the mutual information.

\subsection{Semantic Embedding-and-Retrieval}
Representation learning is one of the greatest concerns of recent machine learning studies~\cite{Bengio2012}.
The primary purpose of representation learning is to build a map that projects a given input to a feature that represents useful information for successive tasks.
A good representation arranges the inputs following their semantic relationships, captures the underlying mechanism of the input, or contains only the information of interest without nuisance factors~\cite{Bengio2012a,Mikolov2013,Ganin2015,Higgins2017a}.

Semantic embedding is a supervised representation learning task~\cite{Weston2010,Frome2013,Kiros2014,Karpathy2015,Zhang2017p,Gu2017,Faghri2018}.
Its purpose is to arrange samples so that similar samples are close to each other in the embedded space.
Typically, a dataset is divided into many small groups of similar samples (e.g., a group of an image and five captions~\cite{Lin2014}).
Given a sample (called a query), the DNN is expected to find similar samples according to their distances in the embedded space.
Form this viewpoint, the semantic embedding is a regression task without identifiability.
However, the performance of the semantic embedding has often been evaluated by the retrieval task, whose purpose is to find the most similar sample to the query from candidates (called targets).
When regarding the given sample as a representative point of a class, the retrieval task becomes a classification task.
In fact, the semantic embedding has been employed for one-shot classification and extreme multi-class classification~\cite{Jain2016,Zhang2017p,Chen2019b}.

\section{Uncertainty Measures for Embedding-and-Retrieval}\label{sec:methods}
\subsection{Feature Uncertainty for Embedding}
For the semantic embedding, a DNN outputs a feature vector $y$ given a sample $x$.
Given a similarity ranking of samples or groups of samples, the DNN is trained to embed a sample so that it is close to similar samples in the embedded space.
Specifically, the distance from a query sample $x_q$ to a designated target $x_p$ (called a positive target) should be closer than that to another target $x_n$ (called a negative target).
The 0-1 loss for this objective is $I(s(y_q,y_p)>s(y_q,y_n))$, where $s(\cdot,\cdot)$ is a similarity function and $I(cond)$ is the indicator function~\cite{Weston2010}.
Because it is difficult to optimize the 0-1 loss, the following hinge rank loss has been widely used~\cite{Rosasco2004,Weston2010,Frome2013,Kiros2014,Karpathy2015,Zhang2017p,Gu2017,Faghri2018}.
\begin{equation}
  \mathcal L_{emb}(x_q,x_p,x_n)=|s(y_q,y_n)-s(y_q,y_p)+m|_+,
\end{equation}
where $|\cdot|_+$ is the positive part and $m>0$ is a margin parameter.
The similarity function $s(\cdot,\cdot)$ can be the negative Euclidean distance, inner product, or cosine similarity in the embedded space.
Note that recent studies have preferred the cosine similarity, which suppresses an excessive similarity and distance~\cite{Gu2017,Faghri2018}.

When the DNNs are BNNs, the similarity function $s(\cdot,\cdot)$ depends on the drawn parameter $w$ and the hinge rank loss is averaged over the parameter posterior $q(w|D)$.
The epistemic uncertainty of a sample $x$ is defined as the variance of the feature vector $y$ in the embedded space, i.e.,
\begin{equation}
  \begin{split}
    \mathcal U_{f}(x)&=Var_{q(w|D)}[y]\\
    &=Var_{q(w|D)}[DNN(x,w)].
  \end{split}
\end{equation}
We call this uncertainty the \emph{feature uncertainty}, hereafter, where the variance is summed over the elements in the feature vector $y$.
We can average the embedded feature $y$ over the parameter posterior $q(w|D)$ as the model averaging; we call this the model averaging over feature uncertainty.

\subsection{Posterior Uncertainty for Retrieval}
For the classification task, the DNN outputs an intermediate feature $y$ given a sample $x$~\cite{Miyato2015,Zhang2017c,Takahashi2018a,Gal2016,Kendall2017a,Kendall2017b,Leibig2016,Xiao2018a}.
Then, a fully-connected (FC) layer and the softmax function are applied to the feature $y$, resulting in a posterior probability $p(c_j|x)$ that the sample $x$ belongs to the class $c_j$ indexed by $j$ as follows:
\begin{equation}
  \begin{split}
    p(c_j|y,\{w_k\})
    &=softmax(y,\{w_k\})_j\\
    &=\frac{\exp((w_j\cdot y)/T)}{\sum_k \exp((w_k\cdot y)/T)},
  \end{split}
\end{equation}
where $\{w_k\}$ is the parameter of the FC layer and $T$ is the temperature parameter.
The bias term is omitted for simplicity.
Given a target posterior $q(c_j|x)$, the objective function of the classification task is typically the cross-entropy loss:
\begin{equation}\textstyle
    \mathcal L_{cls}(x,w,\{w_k\})=\sum_j - q(c_j|x)\log p(c_j|y,\{w_k\}).
\end{equation}
This loss is also called the logistic loss for binary classification~\cite{Rosasco2004}.
The mutual information $I(c;w,\{w_k\}|x,D)$ between the posterior and the parameter was proposed as the epistemic uncertainty~\cite{Smith2018a}, which is calculated as
\begin{equation}
  \begin{split}
    \mathcal U_{p}(x)
    &=I(c;w,\{w_k\}|x,D)\\
    &=H(\mathbb E_{q(w,\{w_k\}|D)}[p(c|x,w,\{w_k\})])\\
    &\ \ \ \ \ -\mathbb E_{q(w,\{w_k\}|D)}[H(p(c|x,w,\{w_k\}))].
  \end{split}
\end{equation}
The mutual information can be calculated as the difference of the entropy before and after the model averaging.
From another viewpoint, the mutual information is a measure of reduction in the entropy of the parameter $w$ after the label $c$ is given.
If the DNN is already trained with samples similar to the given sample $x$, the parameter $w$ is updated only a little and the mutual information is small, and vice versa.
The mutual information measures how informative a new sample $x$ and label $c$ pair is for the DNN.

The cross-entropy loss is commonly used by embedding-and-retrieval systems of person images (i.e., person re-identification systems)~\cite{Li2017f}.
These systems classify an image set of the same person as a class and use the extracted features $y$ as the embedded features for retrieval.
In contrast, this loss is much less common for multi-modal embedding-and-retrieval systems, and their results were often inferior to the results with the hinge rank loss~\cite{Li2017f,Zheng2017a}.
For both multi-modal retrieval system and person re-identification system, we have no weight parameter $w_k$ for a new class in the retrieval phase.
Instead, using the similarity, we assume the posterior probability that a target $x_j$ is the best positive target of all targets $\{x_k\}$ for a query $x_q$ in the retrieval phase as
\begin{equation}
  \begin{split}
    p(c_j|x_q,\{x_k\})
    &=\frac{\exp(s(y_q,y_j)/T)}{\sum_k \exp(s(y_q,y_k)/T)},
  \end{split}
\end{equation}
where the binary variable $c_j$ has a value of 1 if $x_j$ is the best positive target for the query $x_q$.
The embedded feature $y_k$ of a target $x_k$ corresponds to the parameter $w_k$ for the ordinary classification.
Given targets $\{x_k\}$, we define the mutual information $\mathcal U_{p}(x)$ as an uncertainty called the \emph{posterior uncertainty}, hereafter.
We emphasize that, unlike the feature uncertainty, the posterior uncertainty is defined with targets unlike the feature uncertainty.
We can average the retrieval posterior $p(c_j|x_q,\{x_k\})$ over the parameter posterior $q(w|D)$ using the model averaging; we call this the model averaging over posterior uncertainty.
For simplicity, we average the feature vectors $\{y_k\}$ of the targets $\{x_k\}$ before calculating the retrieval posterior and evaluate the posterior uncertainty only for queries.
We have empirically found that this simplification does not influence the performance and results much.
In addition, we empirically confirmed that the variance of the retrieval posterior demonstrated the same tendency as that of the mutual information in all experiments as shown in~\cite{Smith2018a}, and hence, we omitted the results of the variance of the retrieval posterior in this paper.

Note that the similarity function $s(\cdot,\cdot)$ is not necessarily the inner product but typically the cosine similarity, which is bounded in the range $[-1,1]$.
As a result, the posterior probability is never close to 0.0 or 1.0 when $T=1.0$.

\section{Experiments and Results}\label{sec:results}
\subsection{Experimental Settings}
For the image-caption retrieval task, we employed the typical DNN architecture, VSE++~\cite{Faghri2018}.
We used the source code provided by the original authors\footnote{https://github.com/fartashf/vsepp} and the original experimental settings unless otherwise stated.
VSE++ has an image encoder and text encoder.
For image encoding, we used VGG19~\cite{Simonyan2015} pretrained using the ImageNet dataset~\cite{Deng2009}.
We removed the final FC layer for classification and added a new FC layer for embedding.
The dimension number of the embedded space was 1024.
Dropout~\cite{Srivastava2014} with a keep probability $p=0.5$ was already applied before each FC layer.
We resized each input image so that the smaller edge was $256$ and cropped it to a $224\times224$ region randomly in the training phase.
In the retrieval phase, we cropped the center region.
For text encoding, we used a GRU-based text encoder, which is a kind of recurrent neural network~\cite{Cho2014}.
Each word was expressed as a one-hot coded vector and projected to a word embedded space using an affine transformation.
The dimension number of the word embedded space was 300.
Then, a GRU network read the embedded words in a sentence sequentially outputted a vector, which was used as an embedded feature.
In addition to the original experimental settings, we applied dropout with a keep probability $p=0.9$ to the word embedded space to obtain stochasticity.
In the training phase, we minimized the hinge rank loss only with the closest negative target and discarded the losses with other negative targets.
Specifically,
\begin{equation}
  \mathcal L_{emb}(x_q,x_p,\{x_n\})=\max_n |s(y_q,y_n)-s(y_q,y_p)+m|_+,
\end{equation}
where $\{x_n\}$ denotes negative targets in a mini-batch and the margin $m$ was set to 0.2.
The similarity function $s(\cdot,\cdot)$ was the cosine similarity.

\begin{figure*}[t]\centering
  \tabcolsep=0mm
  \begin{tabular}{cccc}
  \includegraphics[scale=0.283]{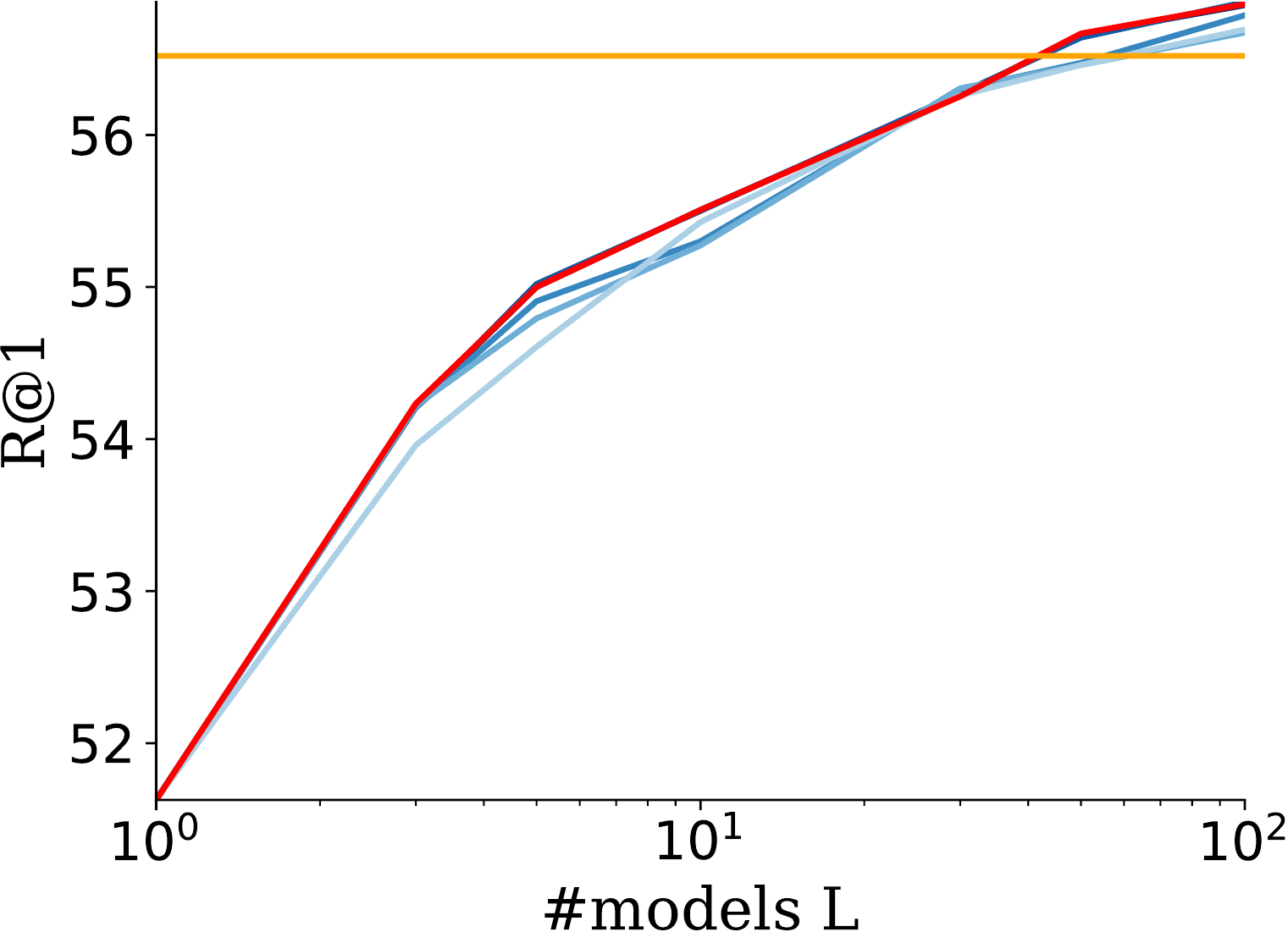}&
  \includegraphics[scale=0.283]{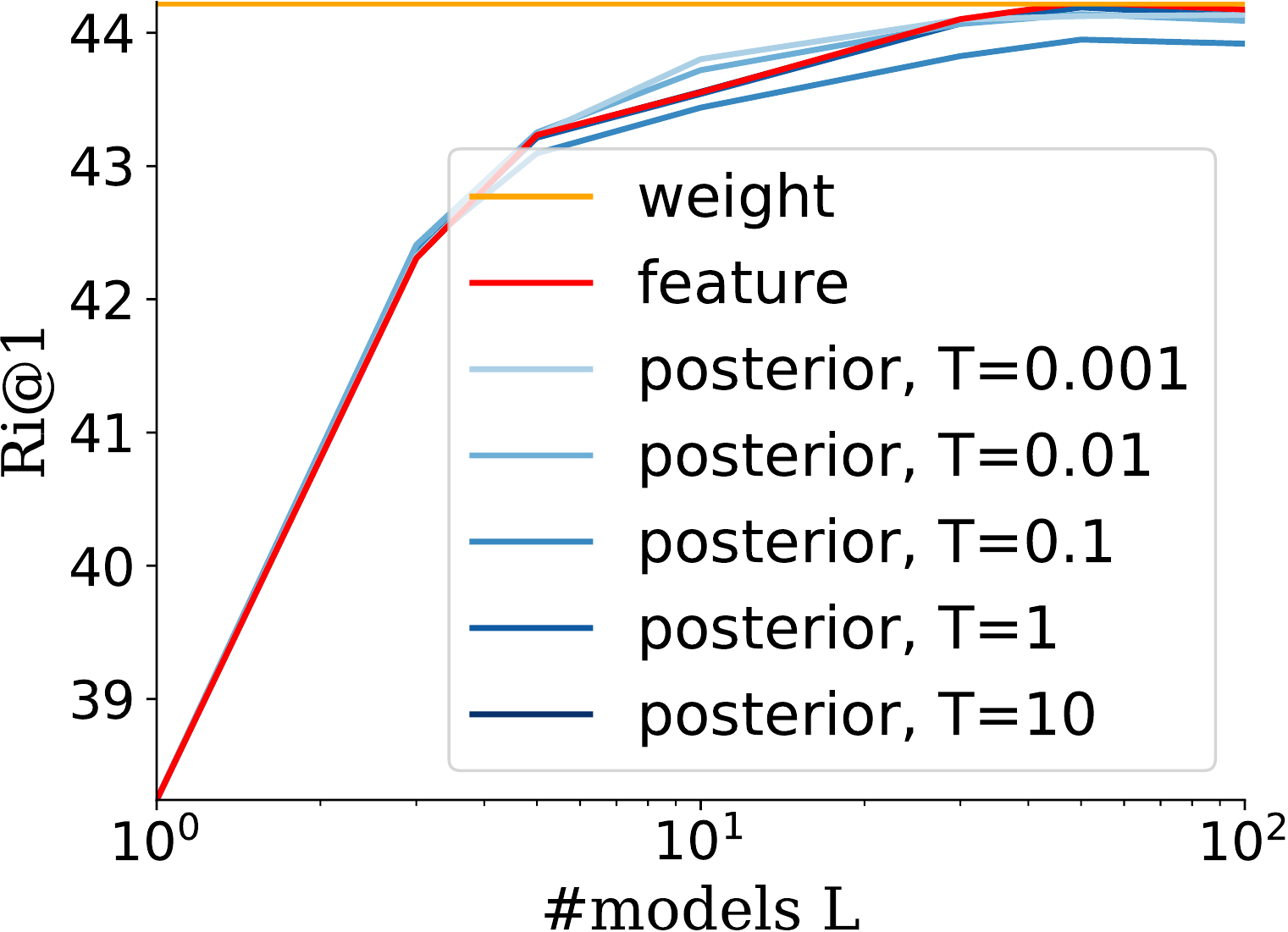}&
  \includegraphics[scale=0.283]{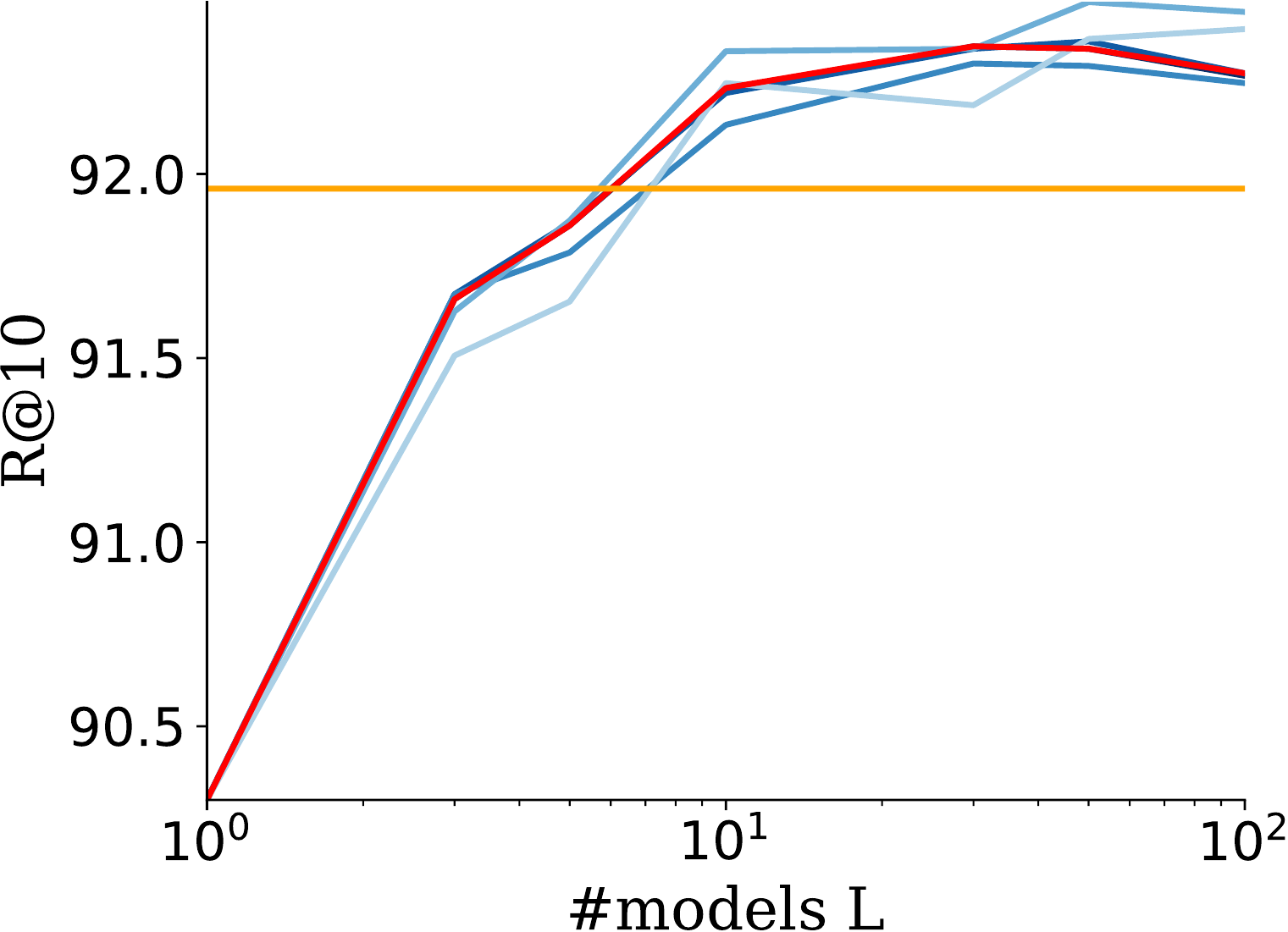}&
  \includegraphics[scale=0.283]{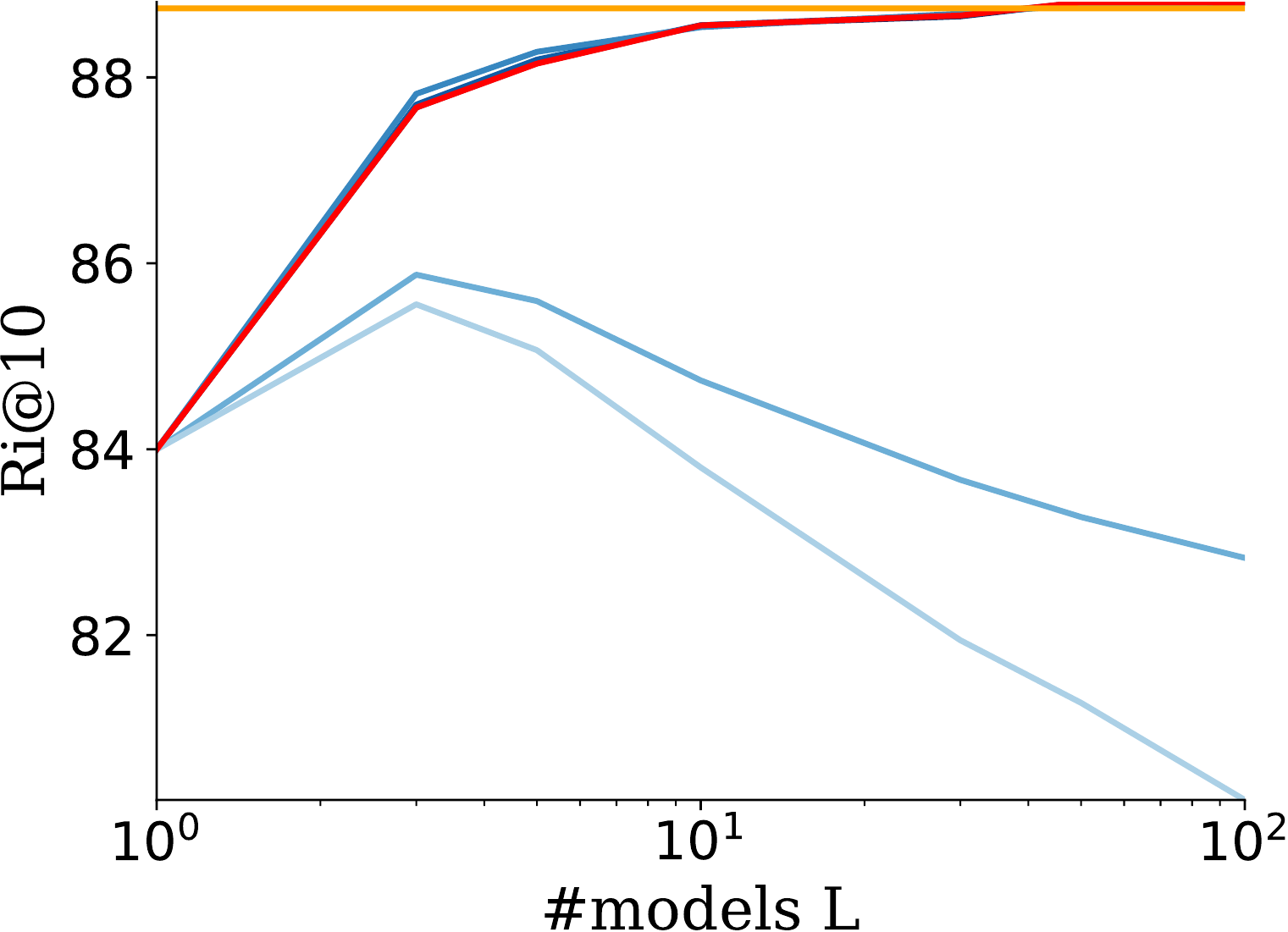}\\
(a)&(b)&(c)&(d)\\[-2mm]
  \end{tabular}
\caption{Retrieval performances of VSE++ with VGG19 using ms coco dataset.}
  \label{fig:ensemble_mscoco}
\end{figure*}

\begin{table*}[t]\centering
  \caption{Retrieval performances}
  \small
  \tabcolsep=1.5mm
  \begin{tabular}{@{}lccccccccccc@{}}
    \toprule
    & \multicolumn{2}{c}{\textbf{Dataset}}  & & \multicolumn{4}{c}{\textbf{Caption Retrieval}} & \multicolumn{4}{c}{\textbf{Image Retrieval}}\\
    \cmidrule(lr){2-3}\cmidrule(lr){5-8}\cmidrule(lr){9-12}
    \textbf{Model} & \textbf{Training} & \textbf{Testing} & \textbf{Averaging} & \textbf{R@1}$\uparrow$ & \textbf{R@5}$\uparrow$ & \textbf{R@10}$\uparrow$ & \textbf{Med r}$\downarrow$ & \textbf{Ri@1}$\uparrow$ & \textbf{Ri@5}$\uparrow$ & \textbf{Ri@10}$\uparrow$ & \textbf{Med r}$\downarrow$\\
    \midrule
    (1) VSE++ with VGG19  & MS COCO   & MS COCO   & weight    &     56.5  & \bf{84.4} &     92.0  & \bf{1.0} & \bf{44.2} &     78.4  &     88.7  & \bf{2.0} \\
                          &           &           & feature   & \bf{56.9} & \bf{84.4} & \bf{92.3} & \bf{1.0} & \bf{44.2} & \bf{78.5} & \bf{88.8} & \bf{2.0} \\
                          &           &           & posterior & \bf{56.9} & \bf{84.4} & \bf{92.3} & \bf{1.0} & \bf{44.2} & \bf{78.5} & \bf{88.8} & \bf{2.0} \\
    \midrule
    (2) VSE++ with VGG19  & Flickr30k & Flickr30k & weight    &     40.1  &     67.5  &     76.7  & \bf{2.0} &     30.0  &     59.8  &     70.3  & \bf{3.3} \\
                          &           &           & feature   & \bf{40.7} & \bf{68.4} & \bf{77.4} & \bf{2.0} & \bf{30.2} & \bf{60.1} & \bf{70.7} & \bf{3.3} \\
                          &           &           & posterior & \bf{40.7} & \bf{68.4} &     77.3  & \bf{2.0} & \bf{30.2} & \bf{60.1} & \bf{70.7} & \bf{3.3} \\
    \midrule
    (3) VSE++ with VGG19  & MS COCO   & Flickr30k & weight    &     35.0  &     61.8  &     73.1  & \bf{3.0} &     25.9  &     52.2  &     63.6  & \bf{5.0} \\
                          &           &           & feature   & \bf{35.8} & \bf{62.6} & \bf{74.0} & \bf{3.0} & \bf{26.1} & \bf{52.3} & \bf{63.9} & \bf{5.0} \\
                          &           &           & posterior & \bf{35.8} & \bf{62.6} & \bf{74.0} & \bf{3.0} & \bf{26.1} & \bf{52.3} &     63.8  & \bf{5.0} \\
    \midrule
    (4) VSE++ with ResNet & MS COCO   & MS COCO   & weight    & \bf{65.3} & \bf{90.1} & \bf{96.0} & \bf{1.0} & \bf{50.4} & \bf{83.3} &     91.6  & \bf{1.3} \\
                          &           &           & feature   &     64.8  &     89.9  &     95.9  & \bf{1.0} & \bf{50.4} & \bf{83.3} & \bf{91.7} & \bf{1.3} \\
                          &           &           & posterior &     64.8  &     89.9  & \bf{96.0} & \bf{1.0} & \bf{50.4} & \bf{83.3} & \bf{91.7} & \bf{1.3} \\
    \midrule
    (5) VSE0 with VGG19   & MS COCO   & MS COCO   & weight    &     50.2  & \bf{81.5} & \bf{90.6} & \bf{1.3} &     38.3  &     74.6  &     87.0  & \bf{2.0} \\
                          &           &           & feature   & \bf{50.4} & \bf{81.5} &     90.5  &     1.4  & \bf{38.8} & \bf{74.8} & \bf{87.1} & \bf{2.0} \\
                          &           &           & posterior & \bf{50.4} & \bf{81.5} &     90.5  &     1.4  & \bf{38.8} & \bf{74.8} & \bf{87.1} & \bf{2.0} \\
    \bottomrule
  \end{tabular}\\
  \flushleft
  \hspace{2cm}$\uparrow$ indicates that a larger value is better and $\downarrow$ indicates that a smaller value is better.\\
  \label{tab:ensemble}
\end{table*}

We evaluated VSE++ on the MS COCO~\cite{Lin2014} and Flickr30k~\cite{Young2014} datasets using the splits used by the original VSE++~\cite{Faghri2018}.
For the MS COCO dataset, we used 113,287 images for training, 5,000 images for validation, and 5,000 images for testing.
We reported the performance averaged over 5-folds of validation/testing images.
For the Flickr30k dataset, we used 30,000 images for training, 1,000 images for validation, and 1,000 images for testing.
Each image has five captions as positive targets.

The VGG19 and GRU network were jointly optimized for 45 epochs using the Adam optimizer~\cite{Kingma2014b} with a batch-size of 128.
The VGG19 except for the final FC layer was freezed for the first 30 epochs and unfreezed for the remaining 15 epochs.
The learning rate was initialized to $2\times10^{-4}$, and then, it was reduced by 0.1 at 15th and 30th epochs.

A typical performance measure of the retrieval is recall at K, which is the fraction of positive targets in the top $K$ candidates.
R@K denotes the performance of caption retrieval based on an image query, and Ri@K denotes that of image retrieval based on a caption query.
The model was evaluated on the sum of R@1, R@5, R@10, Ri@1, Ri@5, and Ri@10 for the validation set after every epoch using the weight averaging, and the best snapshot was selected.
We also reported the performance measure Med r, which is the median rank of the first positive target.

We also evaluated VSE++ with ResNet152~\cite{He2015a} as the image encoder.
ResNet152 has no dropout but employs batch normalization.
We applied the stochastic batch normalization~\cite{Atanov2018} to obtain the stochasticity in the retrieval phase; the stochastic batch normalization learns the distribution of the normalization parameters of batch normalization in the training phase and draws parameters from the distribution in the retrieval phase.
One can regard the ordinary batch normalization in the retrieval phase as the weight averaging.

Moreover, we evaluated VSE0 with VGG19, which employed the inner product as the similarity function and minimized the hinge rank loss averaged over all negative targets in a mini-batch~\cite{Faghri2018} as follows.
\begin{equation}\textstyle
  \mathcal L_{emb}(x_q,x_p,\{x_n\})=\frac{1}{|\{x_n\}|} \sum_n |s(y_q,y_n)-s(y_q,y_p)+m|_+.
\end{equation}
This objective function was formerly used for image-caption retrieval tasks~\cite{Kiros2014}.

\subsection{Retrieval Performance with Model Averaging}
We evaluated the performance on the image-caption retrieval.
First, we focus on VSE++ with VGG19.
We evaluated the model averaging over feature uncertainty and over posterior uncertainty as introduced in Section~\ref{sec:methods}.
For the posterior uncertainty, we adjusted the temperature parameter $T$ from 0.001, 0.01, 0.1, 1.0, and 10.
As a baseline, we evaluated the weight averaging, which averages the stochastic parameters over the posterior distribution.
The weight averaging is a typical way to use dropout and batch normalization in the retrieval phase.
We plotted R@1, Ri@1, R@10, and Ri@10 using the MS COCO dataset with varying the number $L$ of drawn models in Figs.~\ref{fig:ensemble_mscoco} (a)--(d), respectively.
Each result was averaged over 3 runs from scratch.

As the number $L$ of models increases, the retrieval performances obtained with the model averaging improve and finally outperform those of the weight averaging for large $L$ cases.
This result demonstrates that the model averaging works well for image-caption retrieval, as it does for other tasks~\cite{Gal2016,Kendall2017a,Kendall2017b,Leibig2016,Xiao2018a,Taha2019}.
The model averaging over feature uncertainty achieved the best results on average.
The model averaging over posterior uncertainty $T=10$ provided almost the same performances for any number $L$ of models.
A lower temperature improves the performance for R@10 and degrades it for Ri@10.
A lower temperature leads the retrieval posterior $p(c_j|x_q,\{x_k\})$ saturated at 0.0 or 1.0, suppresses the influence of a decision with high confidence, and involves more models in the final decision.
This property potentially improves the performance, but a single incorrect decision can disturb the final decision.
Conversely, with a higher temperature, the retrieval posterior becomes close to linearly with the similarity, reducing the influence of a minority decision and making the final decision more robust.

The improvement of image retrieval is limited compared to that of caption retrieval.
The VGG19 was designed using dropout in the original study~\cite{Simonyan2015} while the GRU network was not.
For the model averaging, the GRU network applied the same stochastic components to a given sample repeatedly.
This behavior leads to excessive variances of the intermediate features and potentially surpasses the improvement by the model averaging.

We summarized the results under other conditions in Table~\ref{tab:ensemble} with the number of models $L=100$ and the temperature $T=10$.
All the results demonstrated that the model averaging yields better performances than the weight averaging, and the posterior uncertainty with the temperature $T=10$ and the feature uncertainty are comparable.
This tendency was common for different datasets (cases (1) and (2)), for dataset shift (case (3)), and for similarity functions (case (1) and (5)).

For ResNet and batch normalization (case (4)), the model averaging using $L=100$ models did not improve the performance.
We found that the model averaging is almost comparable to the weight averaging even in a single model case ($L=1$), gradually improves the performance as $L$ increases, and requires more than $L=300$ models to outperform the weight averaging.
The main purpose of batch normalization is to normalize the moments of intermediate activations over samples, and its stochastic behavior due to the random mini-batch selection is insufficient for obtaining diverse models.

\begin{figure*}[t]\centering
  \tabcolsep=0mm
  \begin{tabular}{cccc}
    \includegraphics[scale=0.283]{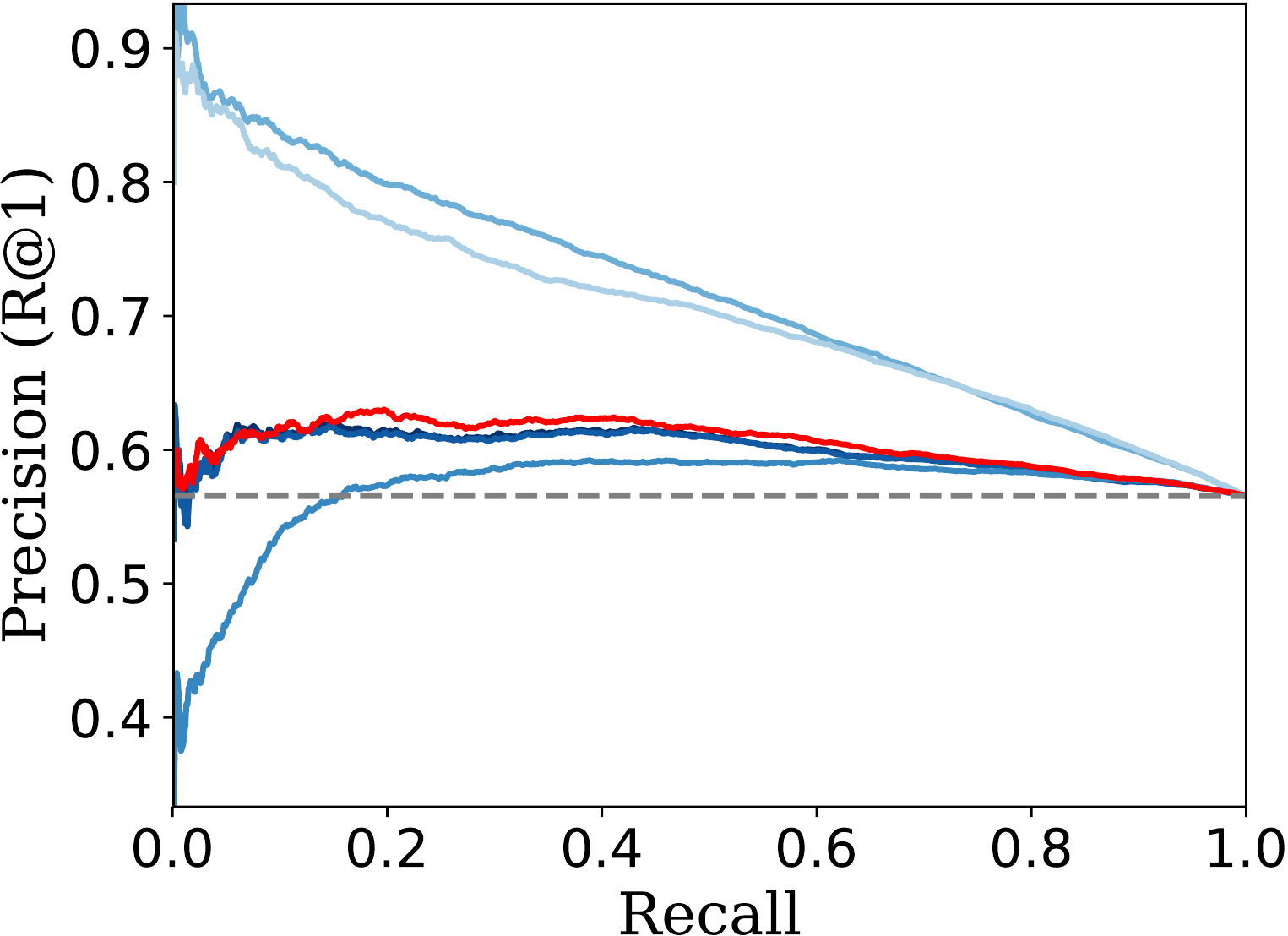}&
    \includegraphics[scale=0.283]{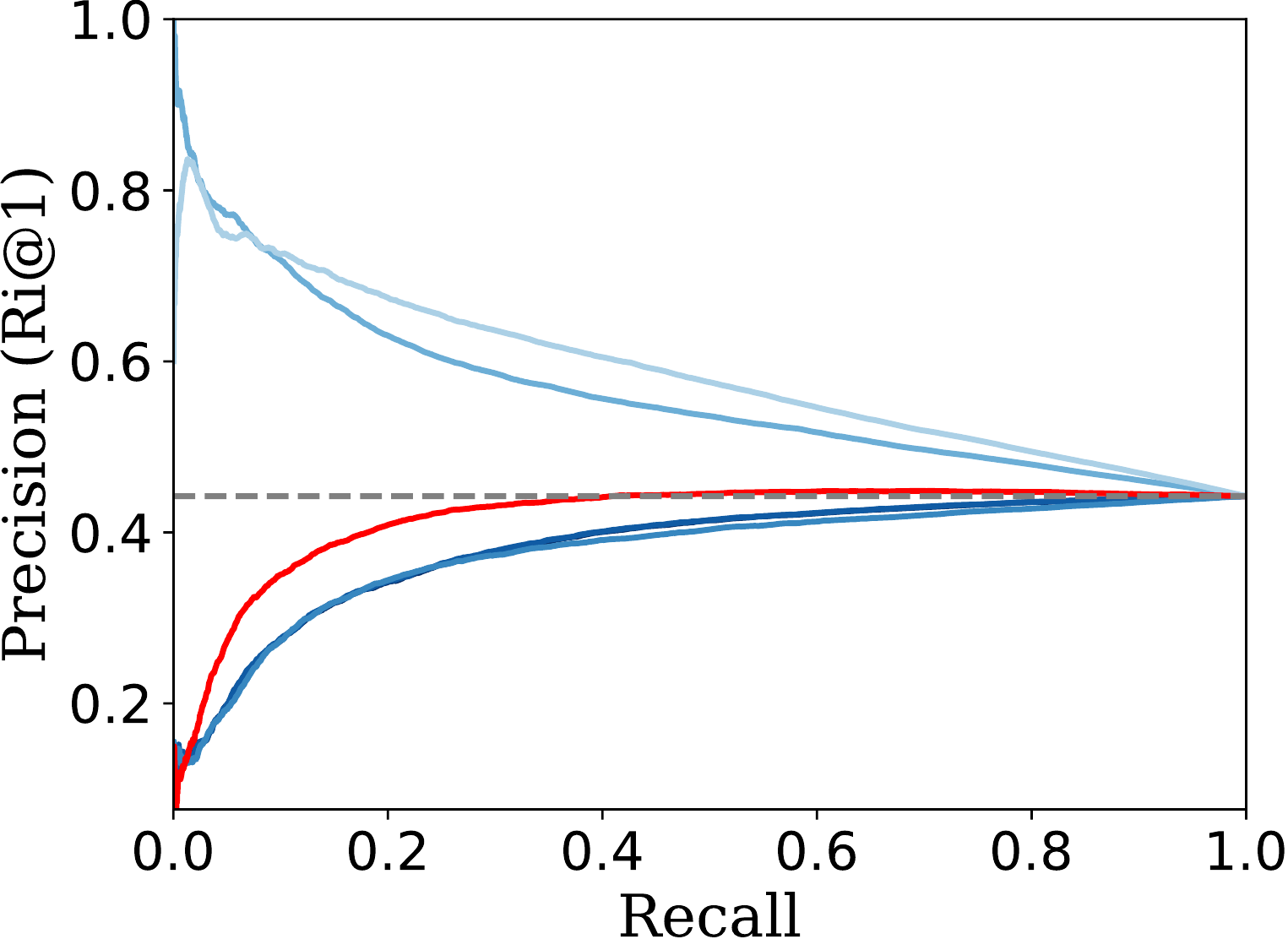}&
    \includegraphics[scale=0.283]{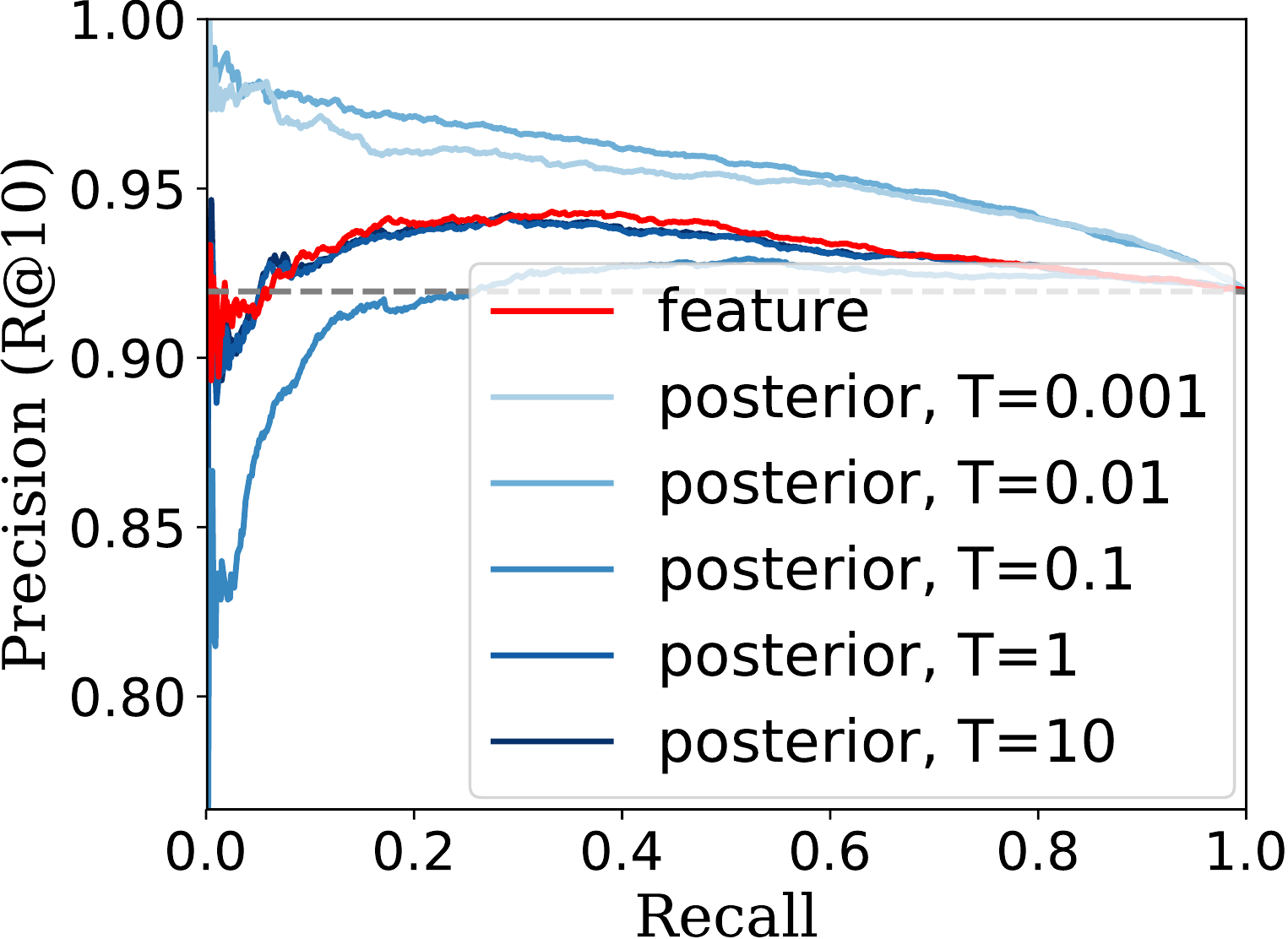}&
    \includegraphics[scale=0.283]{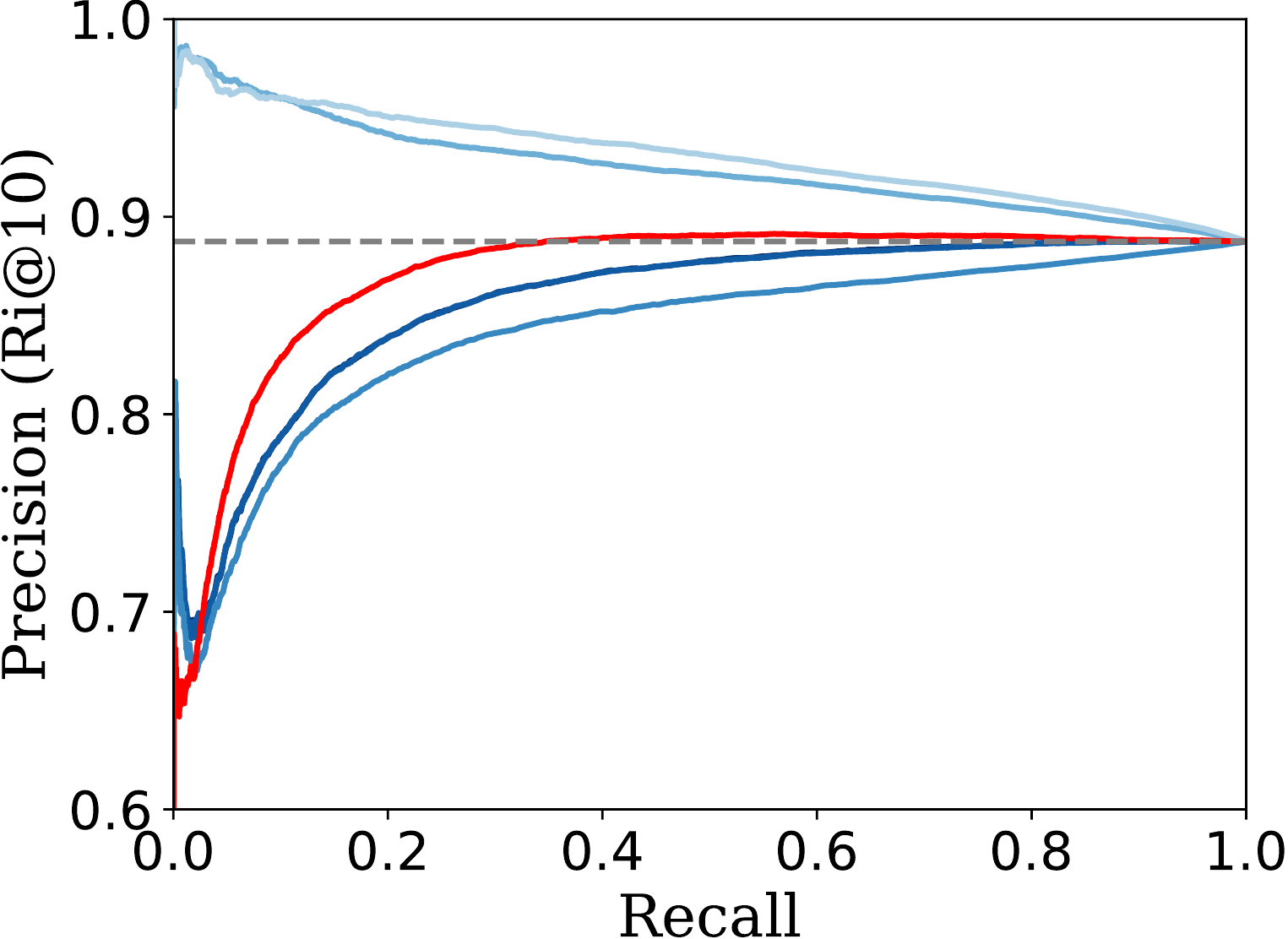}\\
  (a)&(b)&(c)&(d)\\[-2mm]
  \end{tabular}
  \caption{Precision-recall curves of VSE++ with VGG19 on ms coco dataset.}
  \label{fig:prcurve_mscoco}
\end{figure*}

\begin{table*}[t]\centering
  \caption{Area under precision-recall curve (AUPRC) for reliability assessment.}
  \small
  \tabcolsep=1.5mm
  \begin{tabular}{@{}lccccccccccc@{}}
    \toprule
    & \multicolumn{2}{c}{\textbf{Dataset}}  & & \multicolumn{4}{c}{\textbf{Caption Retrieval}} & \multicolumn{4}{c}{\textbf{Image Retrieval}}\\
    \cmidrule(lr){2-3}\cmidrule(lr){5-8}\cmidrule(lr){9-12}
    \textbf{Model} & \textbf{Training} & \textbf{Testing} & \textbf{Uncertainty} & \textbf{R@1}$\uparrow$ & \textbf{R@5}$\uparrow$ & \textbf{R@10}$\uparrow$ & \textbf{Med r}$\downarrow$ & \textbf{Ri@1}$\uparrow$ & \textbf{Ri@5}$\uparrow$ & \textbf{Ri@10}$\uparrow$ & \textbf{Med r}$\downarrow$\\
    \midrule
    (1) VSE++ with VGG19  & MS COCO   & MS COCO   & chance level &     56.5  &     84.4  &     92.0  & \bf{1.0} &     44.2  &     78.4  &     88.7  &     2.0  \\
                          &           &           & feature      &     60.4  &     86.1  &     93.2  &     1.3  &     41.7  &     75.5  &     87.1  &     2.3  \\
                          &           &           & posterior    & \bf{70.4} & \bf{90.3} & \bf{95.2} & \bf{1.0} & \bf{58.6} & \bf{86.0} & \bf{93.1} & \bf{1.2} \\
    \midrule
    (2) VSE++ with VGG19  & Flickr30k & Flickr30k & chance level &     40.1  &     67.5  &     76.7  &     2.0  &     30.0  &     59.8  &     70.3  &     3.3  \\
                          &           &           & feature      &     39.1  &     66.9  &     74.9  &     2.9  &     24.5  &     50.9  &     61.7  &     6.8  \\
                          &           &           & posterior    & \bf{55.5} & \bf{78.8} & \bf{85.8} & \bf{1.3} & \bf{44.2} & \bf{71.9} & \bf{80.3} & \bf{2.1} \\
    \midrule
    (3) VSE++ with VGG19  & MS COCO   & Flickr30k & chance level &     35.0  &     61.8  &     73.1  &     3.0  &     25.9  &     52.2  &     63.6  &     5.0  \\
                          &           &           & feature      &     40.2  &     67.9  &     78.8  &     2.4  &     26.2  &     52.4  &     64.1  &     5.1  \\
                          &           &           & posterior    & \bf{50.4} & \bf{72.4} & \bf{81.8} & \bf{1.7} & \bf{39.4} & \bf{63.8} & \bf{73.0} & \bf{2.8} \\
    \midrule
    (4) VSE++ with ResNet & MS COCO   & MS COCO   & chance level &     65.3  &     90.1  &     96.0  & \bf{1.0} &     50.4  &     83.3  &     91.6  &     1.9  \\
                          &           &           & feature      &     71.6  & \bf{93.5} & \bf{97.7} & \bf{1.0} &     50.1  &     82.4  &     91.1  &     1.9  \\
                          &           &           & posterior    & \bf{75.1} &     93.3  &     97.4  & \bf{1.0} & \bf{65.7} & \bf{90.0} & \bf{95.3} & \bf{1.0} \\
    \midrule
    (5) VSE0 with VGG19   & MS COCO   & MS COCO   & chance level &     50.2  &     81.5  &     90.6  &     1.3  &     38.3  &     74.6  &     87.0  &     2.0  \\
                          &           &           & feature      &     49.3  &     81.5  &     91.1  &     1.6  &     34.6  &     72.9  &     87.0  &     2.4  \\
                          &           &           & posterior    & \bf{63.2} & \bf{88.0} & \bf{94.3} & \bf{1.0} & \bf{50.4} & \bf{81.8} & \bf{91.3} & \bf{1.5} \\
    \bottomrule
  \end{tabular}\\
  \flushleft
  \hspace{2cm}$\uparrow$ indicates that a larger value is better and $\downarrow$ indicates that a smaller value is better.\\
  \label{tab:prcurve}
\end{table*}

\subsection{Reliability Assessment of Image-Caption Retrieval}
We evaluated the relationships between the uncertainty measures and the reliability of the outputs.
A query with a high uncertainty is considered unreliable and tends to lead to mis-retrieval.
Hence, the performance can be improved by rejecting uncertain queries.
There is a trade-off between the fraction of remaining queries (recall) and the performance (precision).
Hence, we report the precision-recall (PR) curves obtained by varying the rejection threshold~\cite{Kendall2017a}.
When an uncertainty measure captures the reliability of retrieval results well, the area under the PR curve (AUPRC) is large.
We evaluated VSE++ with VGG19 using the MS COCO dataset and show the PR curves of R@1, Ri@1, R@10, and Ri@10 in Figs.~\ref{fig:prcurve_mscoco} (a)--(d), respectively.
For fair comparison, the performances were obtained using the weight averaging.
The dotted lines denote the chance levels, where queries are rejected randomly and the precision is unchanged from the case with recall$=1.0$.

As the temperature $T$ decreases, the PR curves obtained from the posterior uncertainty move outward, and the posterior uncertainty with the lowest temperature $T=0.001$ yields the largest AUPRC in most cases.
The posterior uncertainty works well for the reliability assessment of retrieval.

The feature uncertainty yields PR curves lying slightly above the horizontal axes at recall$>0.5$ and rapidly decreasing at a small recall, indicating that VSE++ failed in finding positive targets for queries with small feature uncertainties.
The performance of the feature uncertainty for the reliability assessment is at chance levels or only slightly better than them.

At a higher value of the temperature $T$, the retrieval posterior $p(c_j|x_q,\{x_k\})$ becomes flatter, and any queries and targets are informative; the signal-to-noise ratio of the mutual information is limited.
The same goes for the feature uncertainty; even if queries fluctuate in the embedded space, some of them were always close to their positive targets.
Conversely, at low temperature values, the retrieval posterior $p(c_j|x_q,\{x_k\})$ is saturated at 0.0 or 1.0 and stable with respect to drawn models (i.e., the mutual information is always low) if a query is always close to its targets.
The mutual information is high only when a query and its targets are truly informative and change the parameter $w$ drastically, that is, the case close to mis-retrieval.
Hence, the posterior uncertainty with a low temperature is the best measure of the reliability of the retrieval task.

We summarized the results under other conditions in Table~\ref{tab:prcurve} with the number of models $L=100$ and the temperature $T=0.001$.
The results demonstrate that the posterior uncertainty with the temperature $T=0.001$ worked well as a measure of reliability assessment for all cases; different datasets (cases (1) and (2)), dataset shift (case (3)), different models and stochasticity sources (cases (1) and (4)), and different similarity functions (cases (1) and (5)).
In particular, the posterior uncertainty also provides a reliability measure for ResNet with batch normalization (case (4)); here, the stochastic behavior of batch normalization is sufficient for assessing reliability while being insufficient for the model averaging.
The feature uncertainty was successful only for caption retrieval in cases (1), (3) and (4), and even then its performances were inferior to the posterior uncertainty.
Moreover, the performance is often worse than a chance level.

\begin{figure}[t]\centering
  \includegraphics[scale=0.44]{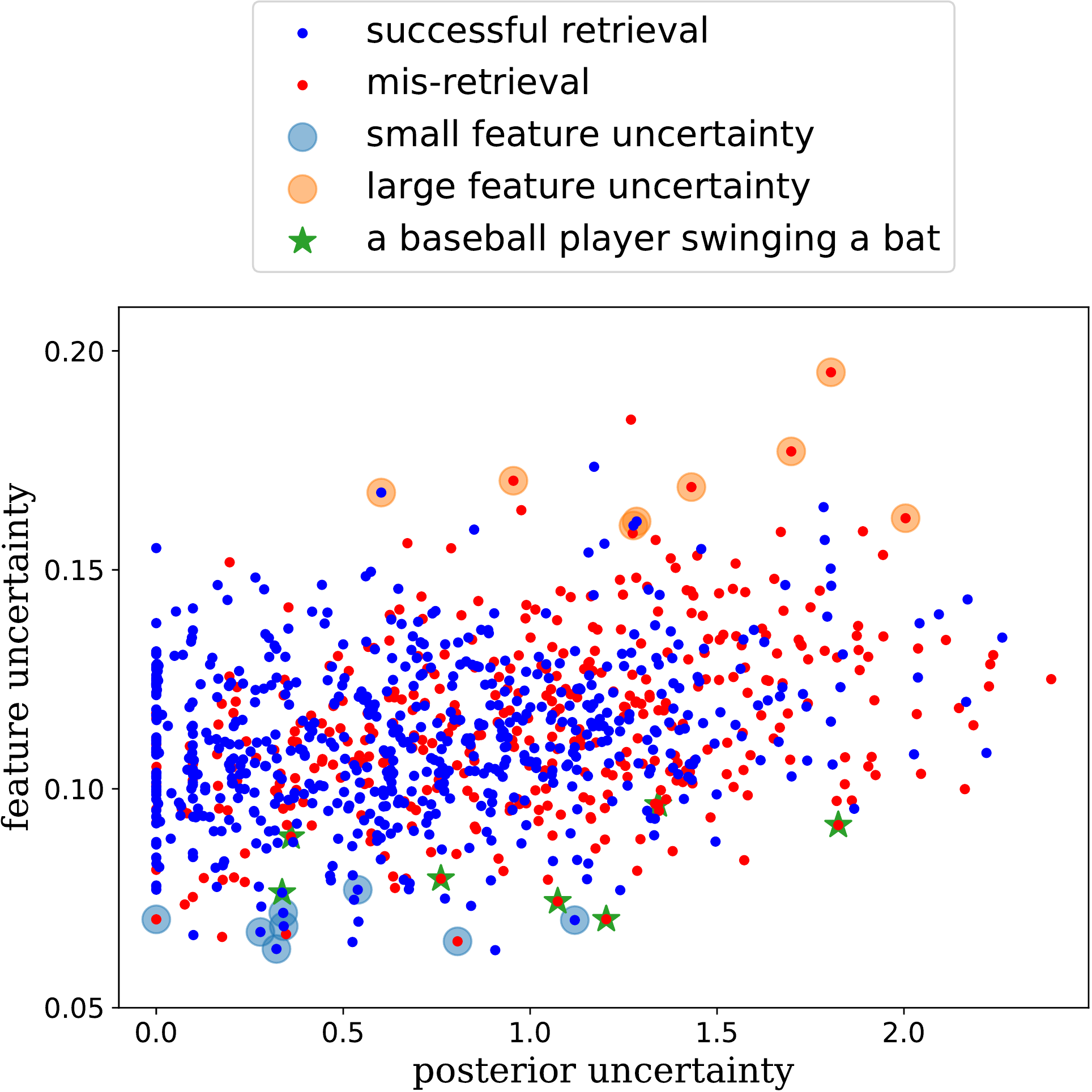}\\[2mm]
  \caption{Scatter plot of uncertainty values for successful and failed caption retrievals on the ms coco dataset.}
  \label{fig:scatter}
\end{figure}
\begin{figure}[t]\centering
  \includegraphics[width=2.2cm, height=2.2cm]{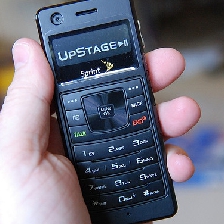}%
  \includegraphics[width=2.2cm, height=2.2cm]{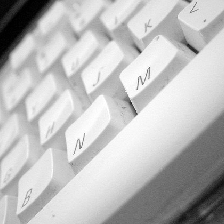}%
  \includegraphics[width=2.2cm, height=2.2cm]{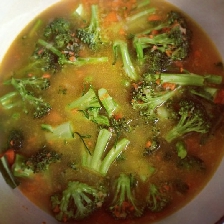}%
  \includegraphics[width=2.2cm, height=2.2cm]{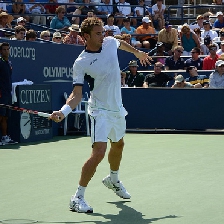}\\
  \includegraphics[width=2.2cm, height=2.2cm]{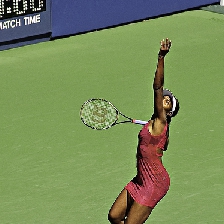}%
  \includegraphics[width=2.2cm, height=2.2cm]{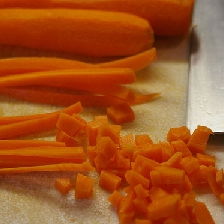}%
  \includegraphics[width=2.2cm, height=2.2cm]{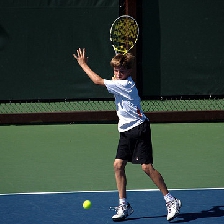}%
  \includegraphics[width=2.2cm, height=2.2cm]{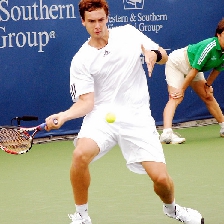}\\
  (a) images of small feature uncertainty\\[3mm]
  \includegraphics[width=2.2cm, height=2.2cm]{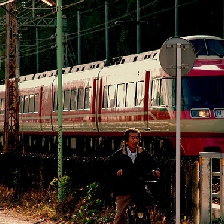}%
  \includegraphics[width=2.2cm, height=2.2cm]{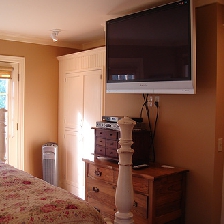}%
  \includegraphics[width=2.2cm, height=2.2cm]{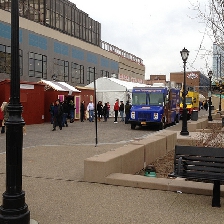}%
  \includegraphics[width=2.2cm, height=2.2cm]{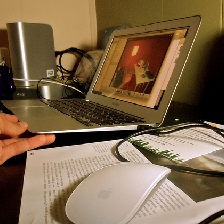}\\
  \includegraphics[width=2.2cm, height=2.2cm]{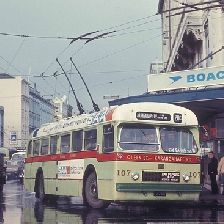}%
  \includegraphics[width=2.2cm, height=2.2cm]{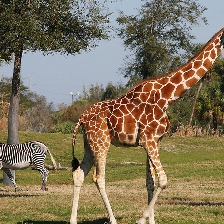}%
  \includegraphics[width=2.2cm, height=2.2cm]{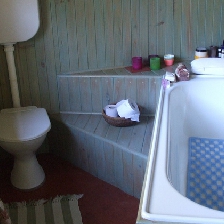}%
  \includegraphics[width=2.2cm, height=2.2cm]{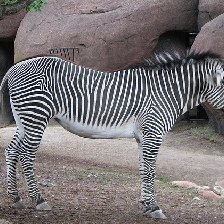}\\
  (b) images of large feature uncertainty\\[3mm]
  \includegraphics[width=2.2cm, height=2.2cm]{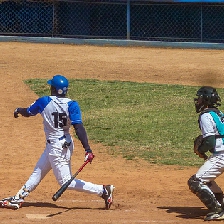}%
  \includegraphics[width=2.2cm, height=2.2cm]{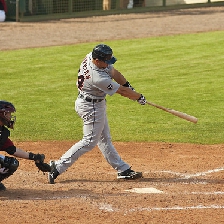}%
  \includegraphics[width=2.2cm, height=2.2cm]{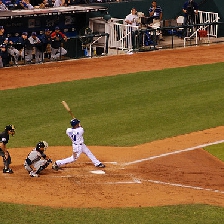}%
  \includegraphics[width=2.2cm, height=2.2cm]{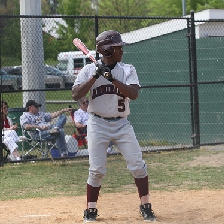}\\
  \includegraphics[width=2.2cm, height=2.2cm]{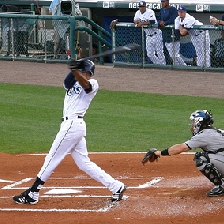}%
  \includegraphics[width=2.2cm, height=2.2cm]{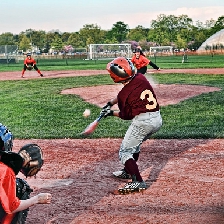}%
  \includegraphics[width=2.2cm, height=2.2cm]{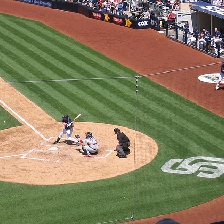}\\
  (c) images depicting a baseball player swinging a bat
  \caption{Example images with small feature uncertainty and large feature uncertainty, and images of a baseball player swinging a bat.}
  \label{fig:exampleimages}
\end{figure}

\subsection{Qualitative Exploration of Large or Small Uncertainties}\label{sec:qualitative}
For a more detailed comparison of the feature uncertainty and the posterior uncertainty, we plotted the uncertainties of images in MS COCO test set in Fig.~\ref{fig:scatter}.
The horizontal axis indicates the posterior uncertainty with the temperature $T=0.001$ and the vertical axis indicates the feature uncertainty.
We also collected example images with large or small feature uncertainties in Fig.~\ref{fig:exampleimages}.
Their uncertainty values are denoted in Fig.~\ref{fig:scatter}.

In Fig.~\ref{fig:exampleimages} (a), example images with small feature uncertainties tend to depict sport players or single objects in close-shots.
In Fig.~\ref{fig:exampleimages} (b), example images with large feature uncertainties tend to depict animals or many objects in long-shots.
This result makes sense because previous studies have demonstrated that the feature uncertainty is large for samples appearing less frequently in the training set~\cite{Gal2016,Atanov2018}.
According to the annotations for object detection, more than 50 \% of the training images depict people and 20 \% are related to sports; the MS COCO dataset is biased to images of people and sports, and the feature uncertainty becomes small for these images.
When an image depicts many objects, the DNN is likely to be unfamiliar with some of the objects, resulting in a large feature uncertainty.
The feature uncertainty successfully captured the biases in the dataset.
However, the feature uncertainty did not evaluate the retrieval performances.

We provide a typical case showing the difference between the feature uncertainty and posterior uncertainty in Fig.~\ref{fig:exampleimages} (c).
The DNN assigns a small feature uncertainty to images depicting a baseball player swinging a bat.
This is because the DNN was trained with many similar samples; according to captions, more than 1 \% of the training samples are related to ``baseball'' and ``swing''.
However, given such a query, the DNN encounters the confusing task of retrieving the best target from many similar targets, and the risk of mis-retrieval increases.
The posterior uncertainty captures this confusion and assigns small or large values by comparing queries with targets (see Fig.~\ref{fig:scatter}).
Conversely, queries about animals and scenery have a limited number of similar queries and relatively distinct from each other.
The DNN discriminates them from each other easily and assigns a small posterior uncertainty even though it is unfamiliar with them and assigns a high feature uncertainty.
The actual reliability of retrieval depends on the relationships with targets more than the population of similar queries, and the posterior uncertainty is suited for the reliability assessment of image-caption retrieval.

\begin{figure}[t]\centering
  \includegraphics[width=5.6cm]{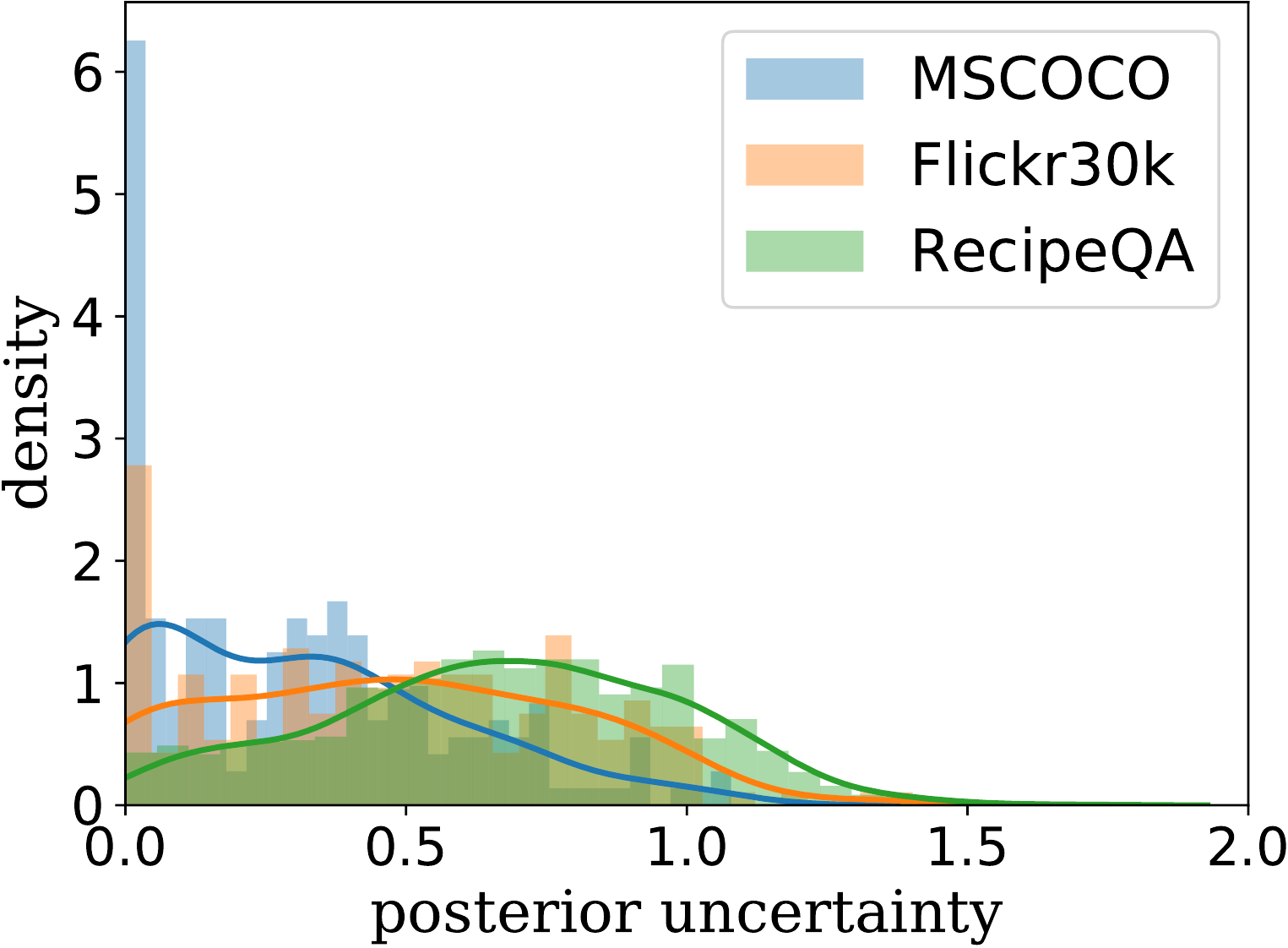}\\[-1mm]
  (a)\\[2mm]
  \includegraphics[width=5.6cm]{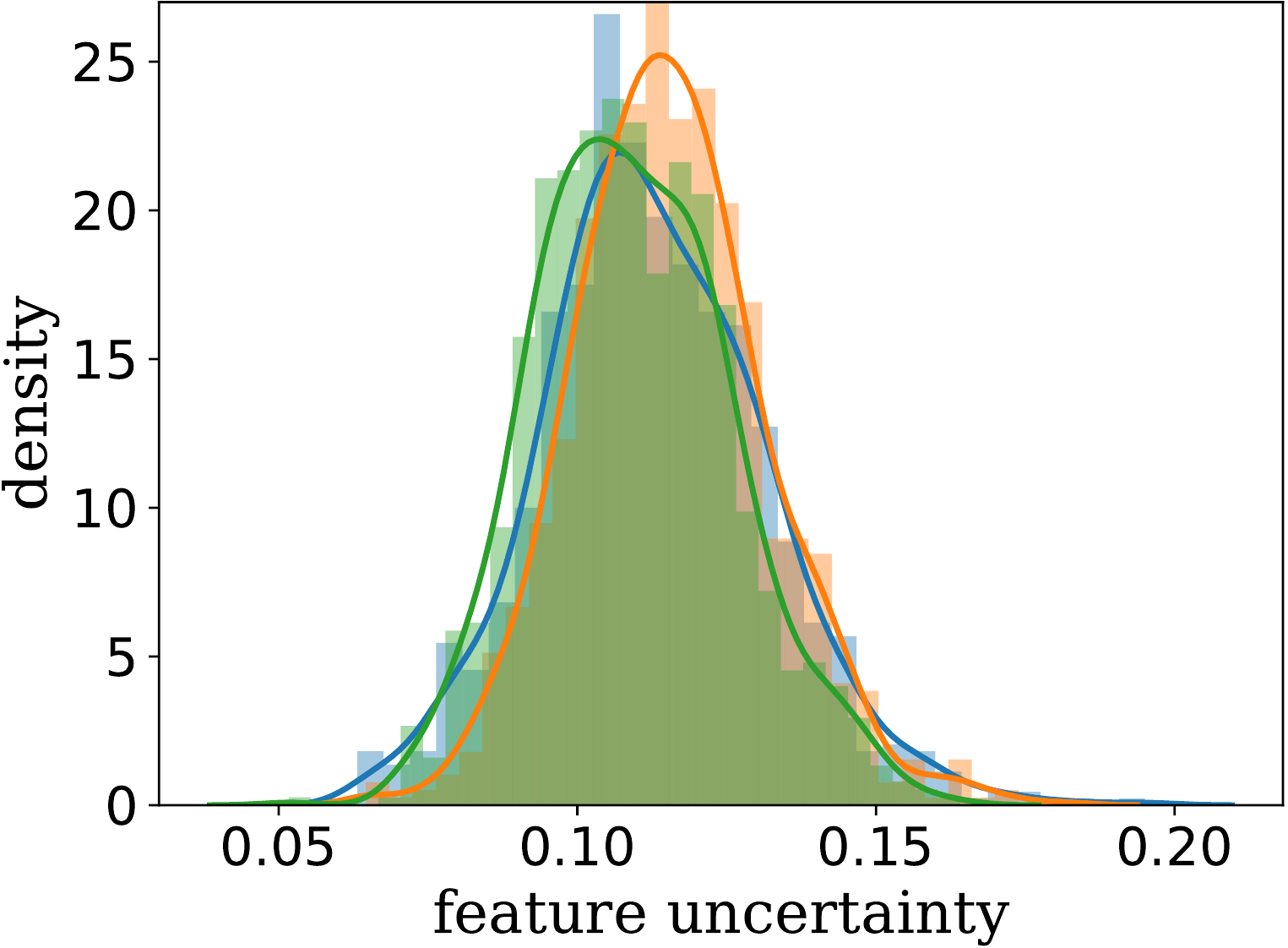}\\[-1mm]
  (b)
  \caption{Distributions of uncertainty measures for caption retrieval under dataset shift.}
  \label{fig:ood}
\end{figure}

\subsection{Reliability Assessment across Datasets}
A trained embedding-and-retrieval system is ideally applicable to any other datasets of images and captions in the same domain.
This situation is called a dataset shift (especially a covariate shift).
For example, as shown in case (3) in Table~\ref{tab:ensemble}, the VSE++ trained using the MS COCO dataset works well for the Flickr30k dataset.
The samples have been gathered and annotated in different pipelines, and each dataset has its own bias.
An uncertainty measure is expected to have a large average value for samples in a different dataset and in an unknown domain, which can be used to detect the dataset shift~\cite{Kendall2017a}.
Here, we assess reliability across datasets.

First, we trained VSE++ with VGG19 using the MS COCO training and validation sets.
Then, we evaluated it using the test sets of MS COCO, Flickr30k, and RecipeQA datasets~\cite{Yagcioglu2018}.
The MS COCO and Flickr30k datasets comprise natural images and their captions, and RecipeQA dataset consists of cooking recipes and corresponding images.
Hence, the MS COCO dataset is further from the RecipeQA dataset than the Flickr30k dataset.
Each recipe in the RecipeQA dataset is associated with several images.
For fair comparison, we randomly chose 1,000 queries and a single target per query.

We plotted the distributions of the uncertainty measures for caption retrieval in Figs.~\ref{fig:ood} (a) and (b).
We set the temperature $T=0.001$ for the posterior uncertainty but we confirmed that the temperature does not influence the tendency.
In Fig.~\ref{fig:ood} (a), the posterior uncertainty is larger on average for the Flickr30k test set than the MS COCO test set and is much larger for the RecipeQA test set.
The posterior uncertainty hence measures the degree of dataset shift.
This result indicates that the posterior uncertainty can be a measure of reliability across datasets.

In Fig.~\ref{fig:ood} (b), the feature uncertainty shows similar distributions for all datasets whereas the performance was highly degraded for Flickr30k dataset (see cases (1) and (3) in Table~\ref{tab:ensemble}).
Moreover, the feature uncertainty of the RecipeQA test set was smaller on average than others.
This result is contrary to previously reported results showing that the outputs for unfamiliar samples have large variances~\cite{Gal2016,Atanov2018}.
As shown in Section~\ref{sec:qualitative}, the feature uncertainty is large for images depicting many objects even when they are obtained in the same domain.
Many images in the RecipeQA depict a few dishes or ingredients and provide smaller feature uncertainty.
Hence, the feature uncertainty does not quantify the difference between the MS COCO, Flickr30k, and RecipeQA datasets and does not work as a reliability measure across datasets.

\section{Conclusion}
This study evaluated two uncertainty measures for image-caption embedding-and-retrieval systems implemented using Bayesian deep learning.
The feature uncertainty, which is designed by considering the embedding as a regression task, improves the retrieval performance by the model averaging consistently.
However, it was found that the feature uncertainty does not assess reliability well in many cases.
The posterior uncertainty, which is designed by considering the retrieval as a classification task, successfully assesses the reliability across samples and across datasets.
These tendencies were common for different datasets, for different DNN architectures, and for different similarity functions.
The qualitative analysis revealed that this difference was caused by the bias in the datasets.
%

%


\end{document}